# MBGDT: Robust Mini-Batch Gradient Descent

Hanming Wang*
hanmingwang2022@u.northwestern.edu
Northwestern University
Evanston, Illinois, USA

Haozheng Luo*
robinluo2022@u.northwestern.edu
Northwestern University
Evanston, Illinois, USA

Yue Wang*
yuewang2023@u.northwestern.edu
Northwestern University
Evanston, Illinois, USA

## ABSTRACT
In high dimensions, most machine learning method perform fragile even there are a little outliers. To address this, we hope to introduce a new method with the base learner, such as Bayesian regression or stochastic gradient descent to solve the problem of the vulnerability in the model. Because the mini-batch gradient descent allows for a more robust convergence than the batch gradient descent, we work a method with the mini-batch gradient descent, called Mini-Batch Gradient Descent with Trimming (MBGDT). Our method show state-of-art performance and have greater robustness than several baselines when we apply our method in designed dataset.

## 1 INTRODUCTION
The existence of outliers bring the machine learning a omnipresent challenge, especially in high dimensions [4]. Most of time, a little number of the outliers will make the machine learning method become fragile, especially in high dimension. How to greater the robustness of the machine learning methods in the high dimension with outliers becomes a controversial problem in the machine learning theory research. There are a lot of research are contributed in the robustness research [3, 5]. However, all of those methods are use the complicated method which influence the efficiency of the training and testing. We hope our method can build based on the basic learner, such as ridge regression or stochastic gradient descent. Mini Batch Gradient Descent (MBGD) is a simple yet effective machine learning model as a linear (and polynomial) regressor and have great robustness than batch gradient descent. However, the naïve MBGD model with squared losses is very sensitive to outliers, making it vulnerable to adversary samples. we designed a method called Mini-Batch Gradient Descent with Trimming (MBGDT). In MBGDT, our group is proposing to add a trimming procedure based on the losses when calculating the gradients to make the MBGD model more robust. We will measure the robustness of the modified model under the $\varepsilon$-contamination model by calculating the Huber Loss on the training sets. Also, we implements some prepossessing methods to make sure MBGDT have higher robustness in training with the data noise. Than we apply our method into designed datasets to compare with several baselines by showing the performance on robustness with random adversaries, adversaries attempting to affect the slopes, and adversaries attempting to affect the bias.

*All authors contributed equally to this research.

In our paper, we introduce the related work of our research and show the achievement and weakness of the recently research on our work. In section 3, we make brief explanation of our model and the technology of trimming the contamination of data and prepossess method in the model. In next section, we will make some experiment with designed database and compare our method performance with baselines to show state-of-art result in our research. In final section, we will discuss some achievement and limitation of our work.

## 2 RELATED WORKS
### 2.1 Gradient Descent Optimization
There are a lot of the technology are developed in the area of the increasing the robustness in the Gradient Descent. From Sebastian Ruder's work [6], the optimization in the Gradient Descent is still a black-box problem in the machine learning research, but there are some popular optimization method work properly on the Gradient Descent. In our paper, we more focus on the optimization on the Mini Batch Gradient Descent (MBGD). The Mini Batch Gradient Descent is a technology which reduce the variance in the estimate of the gradient and have possibility to use several highly optimized matrix optimizations making the algorithm efficient[6]. However, even the MBGD already have a good performance on the training and there are a lot of the optimization work is done in the machine learning work, such as Adam, it still have weak robustness in the machine learning training, especially on the high dimension date training. In our work, we contribute to increase the robustness of MBGD with the database including the random outliers.

### 2.2 Robustness and Contamination
In our paper, we hope to solve the problem in epsilon-contamination model. The Epsilon-contamination model is a model have contamination distribution equation 1

$$(1 - \varepsilon)P_\theta + \varepsilon Q \quad (1)$$

Under this model, data are drawn from equation 1 with probability of $\varepsilon$ to be contaminated by some arbitrary distribution Q. There are a lot of the optimization research focus on robustness and Epsilon-contamination. [1, 3, 5]. There shows a good performance on the contamination dataset and want to increase the robustness of the model. How to increase the robustness in the contamination dataset still a black-box problem. There is still not grantee that all kind of the contamination can be trim by the model with exist method. For example, when the contamination is too close to our data, the model still have huge decrease on the performance. In our research, how to reduce the influence of the contamination and increase the robustness of the model is a huge challenge to us, we hope to work based on the existing research work to MBGD algorithm to make it have larger robustness and can better trim the contamination with the our method. We will make some experiment after that to show our method Mini-Batch Gradient Descent with Trimming shows state-of-art performance in our baselines.

## 3 MODEL
### 3.1 Model Trimming
In the naive Mini-Batch Gradient Descent (MBGD) model, a random batch of samples is selected in each iteration. The gradient of the loss with respect to the weights is calculated, and the weights are updated to the opposite direction of the gradient. This process is repeated until the model converges (little change in the loss after an update) or until the number of iterations reaches a pre-set threshold.

Our modified model adds a step before the weight update. Instead of calculating the gradients based on all samples in a batch, the top epsilon * batch size samples are trimmed, so their effects on the gradients are ignored.

This trimming step is inspired by the robust mean estimator with trimming. The motivation is: If the contamination is close to the true samples, they have little effect on the model. If the contamination is far from the true samples, they are likely to be trimmed more in the addition trimming step we introduced.

The pseudo code of our modified model is presented as algorithm 1:

---
**Algorithm 1:** Procedure Fit

Initialize w;
**while** *not converged and not exceeding max-iter iterations* **do**
  Randomly select $batch_size samples without replacement$;
  Calculate the losses of all the samples in the batch;
  Calculate the gradient of the losses with respect to w,
    ignoring the effects of the $\epsilon \times batch_size$;
  samples with the largest losses;
  Update w;
**end**

---

Furthermore, we replaced the squared loss in the naive model with the Huber loss to weaken the effects of the extreme samples and outliers.

### 3.2 Preprocessor
We also developed 2 different preprocessors to combine or reduce the number of dense subsets of the samples, which are in most cases, adversary.

### 3.3 Kernel prepossessor
The kernel preprocessor is inspired by the kernels in convolutional neural networks. The kernel preprocessor moves a kernel with a pre-set kernel size and stride across the domains of x and y. If the number of samples captured by a kernel exceeds a pre-set threshold (a portion of the total sample size), the samples in the kernel are combined in some ways (e.g. a transformed sample with x and y being the means of those of the original samples captured by the kernel). On the the transformed samples generated by the preprocessor are feed to the model. The pseudo code of the kernel preprocessor is presented as algorithm 2:

---
**Algorithm 2:** Procedure Kernel Prepossessing

**for** *all samples* **do**
  calculate the lowest and highest indices of the kernels that include the sample
**end**
**for** *all kernels* **do**
  combine the samples within the kernel, if and only if the number of samples is larger than the threshold
**end**
Return the combined samples only

---

### 3.4 Clustering prepossessor
Clustering preprocessing is a method to solve the edge contamination cases. The main idea of our preprocessing is to find the edge contamination according to its density and reduce these samples from our training sets. To make this method works, we have two basis assumption. The first is that our contamination is showed in a specific domain with larger density than true samples. The second is that some true sample being trimmed will not effect the performance.

We chose the DBSCAN clustering method as a basis. Different from normal DBSCAN method, here we treated contamination as one clustering and the other as noises to find it out based on density. The process can be describe as follows:

---
**Algorithm 3:** DBSCAN

Input: $D, \epsilon, minsamples$
Output: $D'$ Initialize core sets, k ← 0
**while** $dot \quad i \in D$ **do**
  Find N(i) in distance $\epsilon$
  **if** $|N(i)| \leq minsamples$ **then**
    add i to core set k
    Delete core set k from D
  **else**
    k ← k + 1
    continuing
Delete core set 1

---

We can see from the following figure, the red dots represents the samples after preprocessing and the grey dots are the dots of training samples including the contamination. We can see that the grey dots has been clarified and reduced from the training samples.

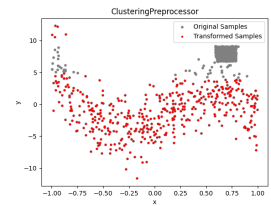

**Figure 1: clustering process**

Compared with the Kmeans clustering, the DBSCAN is based on density and doesn't need initializing the clustering class size. We also take our two basic assumptions into consideration. For the first one of density, we also need to ensure the performance under other uniform contamination. It is ensured when the random uniform contamination won't be more dense than true sample and therefore it won't be regarded as a part of the contamination. For the second assumption of the processed true sample, we have support it on the failure boundary section.

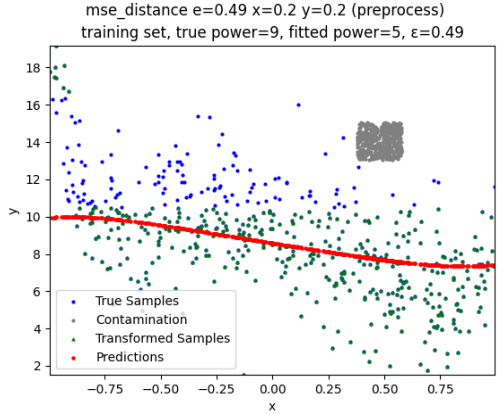

Figure 2: clustering process in testing cases

We can see from the figure above. The figure represents the samples. The grey dots are the contamination, the blue dots are the true samples and the green dots are the training samples after preprocessing.

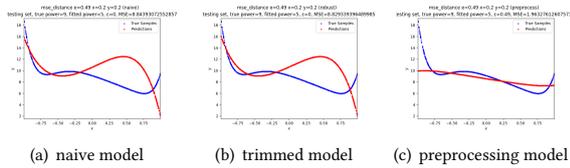

(a) naive model  (b) trimmed model  (c) preprocessing model

Figure 3: Comparison between naive, trimmed and preprocessing

We can see the prediction performance on the above three figures. The preprocessing has obviously narrow the derivation to the true testing samples while the naive model and trimmed model can't. This is the case when the edge contamination is far away from the training set, and for those case that it is near to the training sets we can also expect good performance since the lackness won't effect the performance even is large portion of true samples will be deleted.

And there are also some to-do jobs such as finetuning precess to match the best $\epsilon$ and minisamples. And if there is no edge contamination, the clustering will failed to find the dense part and will delete nearly all the true samples uncorrectly.

## 4 EXPERIMENT
### 4.1 Loss Function and Performance Evaluation

In original MBGD model, the optimizer use the square loss. However, the squared loss has the shortcoming that it has the tendency to be dominated by outliers. As a result, in our model, we replace the square loss with the Huber loss (as equation 2). The Huber loss can have our model has less influence outliers and have a better robustness of the median-unbiased estimator.

When we test the designed dataset to show the robustness of the model, we use the normal MSE model as the evaluation method. The training and testing of the dataset still are a regression behavior. Using the MSE loss can help us see the performance of the regression model works on the designed dataset.

$$L_\delta(a) = \begin{cases} \frac{1}{2}a^2 & |a| \leq \delta \\ \delta(|a| - \frac{1}{2}\delta) & |a| > \delta \end{cases} \quad (2)$$

### 4.2 Contamination Pattern Families

We listed several patterns of possible contamination, and tested all of them with epsilon = 0.49 (except for the no contamination family). We are fitting 9-degree polynomials with 5-degree polynomials to further test the robustness of the models and the performance of the models in high dimensions.

*4.2.1 No contamination.* As shown in figure 4, we tested on the samples without contamination, both with noise and without noise. The results shows that our modified model is virtually the same as the naive model, which means that it doesn't perform worse on the base cases with no contamination.

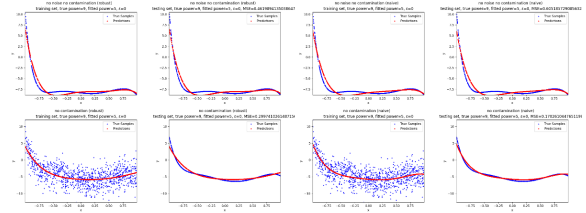

Figure 4: The result of the No Contamination experiment

*4.2.2 Random contamination.* The random contamination family contains the conditions where contamination is uniformly distributed in the domains of x and y. As shown in figure 5, the fitted curve in the naive model is flattened by the contamination, with a mean square error of as high as 6.56.

Our robust model, as shown in figure 6, however, is not heavily affected by the contamination, having a mean square error of only 0.913.

*4.2.3 Parallel Line contamination.* The parallel line contamination family contains the conditions where the adversaries form a parallel line/curve to the true samples. As shown in figure 7, the fitted curve in the naive model is in the middle between the true samples and the contamination, with a mean square error of as high as 994.

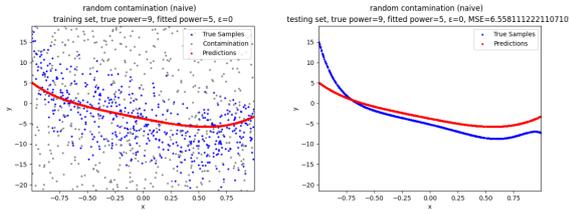

**Figure 5: The result of the Random Contamination experiment (naive model)**

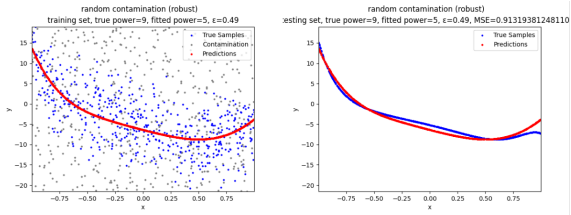

**Figure 6: The result of the Random Contamination experiment (robust model)**

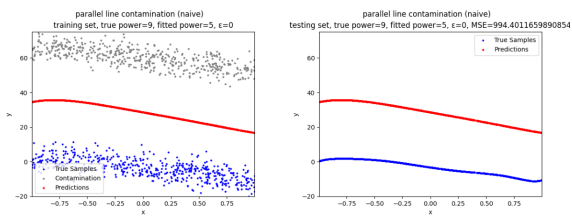

**Figure 7: The result of the Parallel Line Contamination experiment (naive model)**

Our robust model, as shown in figure 8, however, is a close fit to the true samples only, having a mean square error of only 1.29.

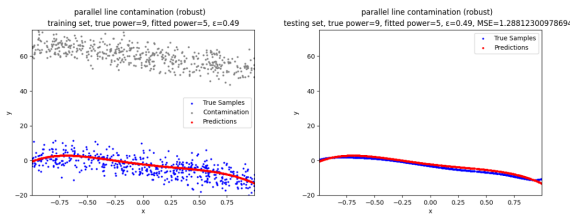

**Figure 8: The result of the Parallel Line Contamination experiment (robust model)**

*4.2.4 Edge contamination.* In our experiment, we shows the four kinds of the edge contamination in our designed dataset, (figure 9). The most basic edge contamination is the condition when there are a huge contamination in the left-top or right-bottom place. Another three edge contamination exists the conditions, if there are one parts of the original (begin, middle, end) data become the contamination and move far away the original data. In our experiment, we want to see the performance of those four kinds of the edge contamination. In figure 9, the left two columns is the training and testing result of the model with trimming, and the right two columns is the training and testing performance of the original MBGD. We can obviously see our model MBGDT bring a huge improvement from original MBGD. The prediction curves closer to the true sample, and the bump are smaller than original model. The huge improvement of MBGDT also show on the loss value (as table 1), the MSE loss of the testing result in MBGDT is much smaller than those of original model MBGD.

| Contamination | MBGDT | MBGD |
|---|---|---|
| Edge Contamination | 26.156 | 692.801 |
| Begin Contamination | 846.498 | 2405.999 |
| End Contamination | 609.331 | 2070.335 |
| Middle Contamination | 156.854 | 1869.965 |

**Table 1: The MSE loss of the Edge Contamination**

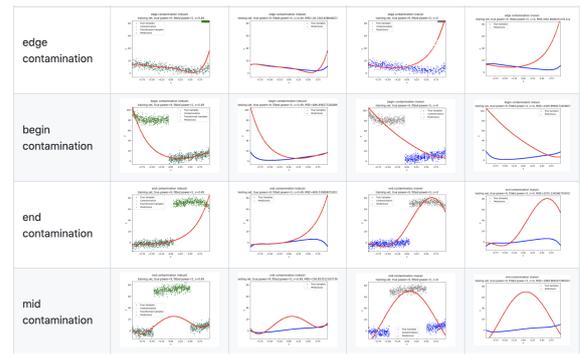

**Figure 9: The result of the Edge Contamination experiment**

### 4.3 Failure Boundaries

To explore the influence of different parameters in the contamination, we did several quantitative test to find the limitations and boundaries of our algorithm. In this section, we mainly have experiments in scale of epsilon, distance and non-uniform distribution of samples. We use mean square error (MSE) to demonstrate the derivation between prediction of trained model and true samples.

*4.3.1 MSE verse epsilon.* Epsilon indicates the size of contamination. When we chose the edge contamination located in the right corner (as the figure 1 shows), we can see that the MSE grows approximately linearly with epsilon (as the figure 2 shows). All the datas are an average for 50 repeated experiments to have a universe statics. In the meantime, our trimmed model has better performance compared with naive mini batch gradient algorithm.

The result is reasonable since the more contamination is, the more derivation it will cause to the trained parameters. And we also find several unexpected behaviors. For example, the naive model will has a sharp drop down where epsilon is around 0.3.

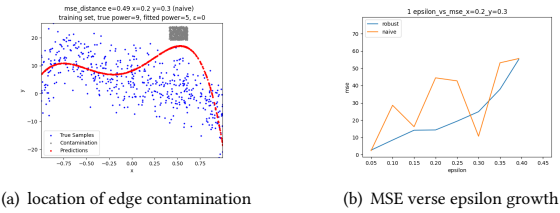

(a) location of edge contamination  (b) MSE verse epsilon growth

Figure 10: MSE verse epsilon

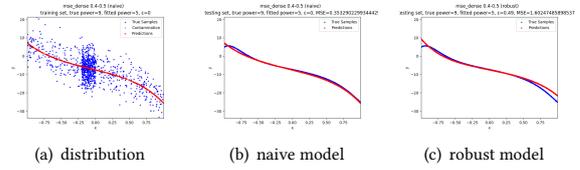

(a) distribution  (b) naive model  (c) robust model

Figure 12: Non-uniform samples in different density

*4.3.2 MSE verse distance.* Since the distance from the edge contamination will influence the loss, it surely will do different effect with the performance. In this experiment, we explore the influence of distance in x-axis and y-axis. To clarify, the distance is showed as a ratio of the distance between the center of contamination divide the length of domain of the axis.

For case 1, the sample distribute more dense in some domain and being sparse in the others. The density sample accounts for more than half of the sample size. We can see the scatter plot for the sample distributions.

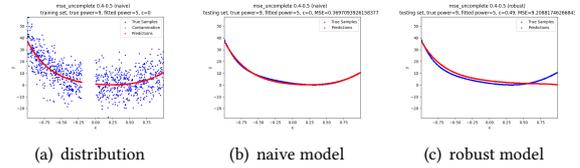

(a) distribution  (b) naive model  (c) robust model

Figure 13: Non-uniform samples in incomplete cases

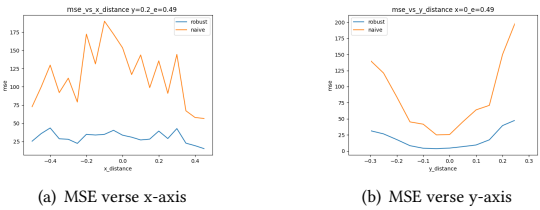

(a) MSE verse x-axis  (b) MSE verse y-axis

Figure 11: MSE verse distance

As the figure shows, the distance in x-axis won't effect the influence. The slight wave maintain in a line level. This is because that among the overall performance in repeated times, the x distance is just a probability problem and has the same influence to the model training.

In y-axis, the MSE grows more than linearly with the distance grows. And compared between naive model and our trimmed model, we can see our model not only reduced the mean square error but also decrease the growth rate for the distance. That represent both robustness and consistency of our method.

*4.3.3 MSE verse non-uniformed samples.* Since the mini-batch is drawn from the training set uniformly and our trim is based on the loss of those samples, we need to make sure the different density in training sample won't effect the performance of our model. To make sure that trim won't delete the true samples and lead to derivation, we did experiments on non-uniform sample, mainly targeted at two different cases.

For case 2, the sample has lackness on some of the domains as the figure shows. The lackness accounts for 10 percent of the training sample size. This is the parameter of the x-width of our testing edge contamination.

For both cases, the prediction of naive model and robust model keep good performance without contamination. This means that our trimmed model won't trim useful true samples which will do harm to the training parameters. And this will support our clustering preprocessing method in later section such that trimmed samples won't effect the performance.

## 5 DISCUSSION AND FUTURE WORK

This paper presented our model MBGDT have a huge improvement than the naïve MBGD model. We obverse the MSE score is reduce a lot from our MBGDT model with our baselines in our designed dataset. Also, the model prediction curve closer to the actual curve. It shows that our model with trimming brings a huge help in the robustness of the model. Also, we find the prepossessing method also increase the robustness of our model. However, there are also some limitation of our MBGDT model. The high-dimension model even will have a huge influence when there are outliers.

Future work will revolve around fine-tuned prepossessing. we will find the suitable e, mini-samples for DBSCAN model and suitable Kernel size, Strides, Threshold for kernel prepossessing. Also, we will also implement Scheffé estimate [2] to show whether it can help increase of the robustness in our designed dataset.